\pdfoutput=1

\documentclass[11pt]{article}

\usepackage[final]{acl}

\usepackage{times}
\usepackage{latexsym}

\usepackage[T1]{fontenc}

\usepackage[utf8]{inputenc}

\usepackage{microtype}

\usepackage{inconsolata}

\usepackage{graphicx}

\usepackage{multirow}
\usepackage{booktabs}
\usepackage{listings}
\usepackage{kotex}
\usepackage{amssymb}
\usepackage{amsmath}

\lstdefinelanguage{dict}{
    breaklines=true,
    breakatwhitespace=true,
    basicstyle=\ttfamily\small,
    upquote=true,
    breakindent=0pt,
}
\usepackage{arydshln}
\makeatletter
\def\adl@drawiv#1#2#3{%
        \hskip.5\tabcolsep
        \xleaders#3{#2.5\@tempdimb #1{1}#2.5\@tempdimb}%
                #2\z@ plus1fil minus1fil\relax
        \hskip.5\tabcolsep}
\newcommand{\cdashlinelr}[1]{%
  \noalign{\vskip 1pt
           \global\let\@dashdrawstore\adl@draw
           \global\let\adl@draw\adl@drawiv}
  \cdashline{#1}[.4pt/2pt]
  \noalign{\global\let\adl@draw\@dashdrawstore
           \vskip 1pt}}
\makeatother

\title{CoME: An Unlearning-based Approach to Conflict-free Model Editing}

\author{Dahyun Jung \qquad Jaehyung Seo \qquad Jaewook Lee \\ \textbf{Chanjun Park}\thanks{Corresponding author.} \qquad \textbf{Heuiseok Lim}\footnotemark[1] \\
  Korea University\\
  \texttt{\{dhaabb55,seojae777,jaewook133,bcj1210,limhseok\}@korea.ac.kr}}

\begin{document}
\maketitle
\begin{abstract}
Large language models (LLMs) often retain outdated or incorrect information from pre-training, which undermines their reliability. While model editing methods have been developed to address such errors without full re-training, they frequently suffer from knowledge conflicts, where outdated information interferes with new knowledge. In this work, we propose Conflict-free Model Editing (CoME), a novel framework that enhances the accuracy of knowledge updates in LLMs by selectively removing outdated knowledge. CoME leverages unlearning to mitigate knowledge interference, allowing new information to be integrated without compromising relevant linguistic features. Through experiments on GPT-J and LLaMA-3 using Counterfact and ZsRE datasets, we demonstrate that CoME improves both editing accuracy and model reliability when applied to existing editing methods. Our results highlight that the targeted removal of outdated knowledge is crucial for enhancing model editing effectiveness and maintaining the model's generative performance. Our code is available at \url{https://github.com/ekgus9/COME}.
\end{abstract}

\section{Introduction}
Large language models (LLMs) encode vast amounts of knowledge during pre-training, enabling them to perform effectively across a wide range of natural language processing (NLP) tasks~\cite{hao2021selfattention,cao2021knowledgeableeducatedguessrevisiting,jiang2023mistral,hernandez2023inspecting,haviv2023understanding,openai2023gpt4}. However, LLMs often incorporate outdated, incorrect, or biased information learned from training data, which can directly affect the reliability of their outputs~\cite{hase2021language,pagnoni2021understanding,Ji_2023,mousavi2024llm}. Such issues may lead to unexpected results or undesirable biases in the generated responses.

There is a growing need for research aimed at correcting erroneous knowledge in LLMs or injecting new knowledge while preserving the general performance of the models. 
Recent studies explore model editing, which offers the potential to modify a model's knowledge without requiring full re-training~\cite{mitchell2022memorybased,wang2023knowledge,yao2023editinglargelanguagemodels,pinter2023emptyingoceanspoonedit,zhang2024comprehensive}. Model editing enables the integration of new information into a model through minimal parameter updates while preserving its existing knowledge. This is particularly useful for correcting errors introduced by flawed data or incorporating new knowledge while selectively updating only the necessary parts of the model.

Existing model editing methods primarily focus on identifying and modifying the parameters where knowledge is stored in order to update the model~\cite{dai2022knowledgeneuronspretrainedtransformers,meng2023massediting,hu2024wilkewiselayerknowledgeeditor,chen2024largelanguagemodelbias,sharma2024locatingeditingfactualassociations,wang2024detoxifyinglargelanguagemodels}. These approaches allow the model to retain learned information efficiently while updating specific knowledge. However, when generating responses based on newly integrated knowledge, the model may encounter conflicts between the new and outdated knowledge, leading to degraded performance~\cite{li2024unveilingpitfallsknowledgeediting}. \citet{ni2024forgettinglearningutilizingparametric} propose a full fine-tuning-based approach that first performs forgetting outdated knowledge before editing the model with new information. However, fine-tuning-based editing is susceptible to overfitting~\cite{decao2021editing}, and updating all layers incurs significant memory overhead. Additionally, the gap between the unlearning and editing stages may lead to unintended knowledge distortions.

\begin{figure*}[]
\centering
\includegraphics[width=0.91\linewidth]{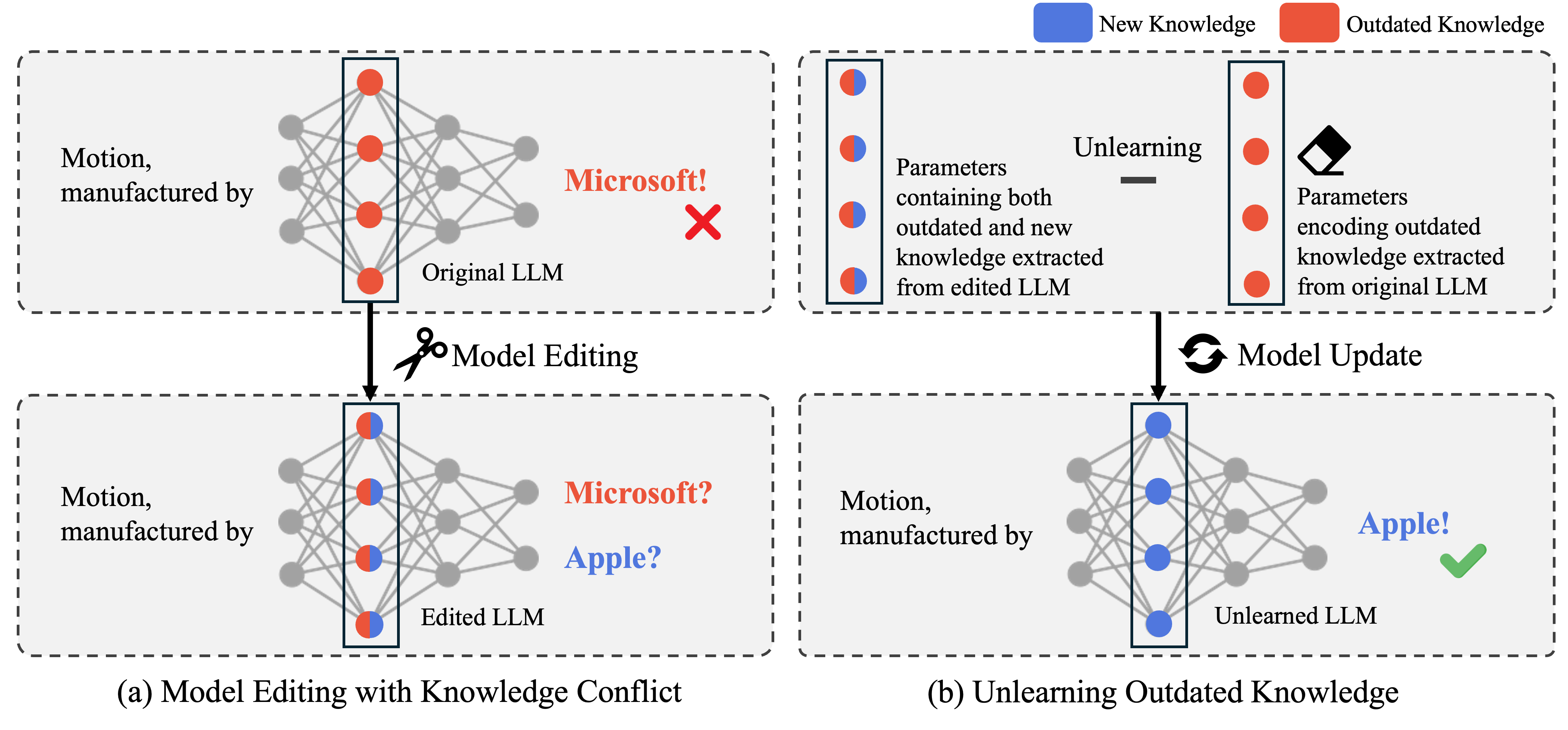}
\caption{The overall framework of CoME. (a) Existing model editing creates a situation where outdated and new knowledge coexist, and (b) we resolve this issue by unlearning the parameters representing the outdated knowledge.}
\label{fig:main}
\end{figure*}



To address these issues, we propose \textbf{Conflict-free Model Editing (CoME)}, which selectively removes outdated knowledge while simultaneously updating the model with new knowledge. This process mirrors the way the human brain refines its understanding—when we learn new information, the brain selectively weakens outdated or conflicting memories to avoid cognitive interference and confusion~\cite{geiselman1983disrupted,bjork1996continuing,wixted2004psychology,alves2017retroactive,kliegl2021mechanisms}. In a similar manner, CoME identifies parameters associated with outdated knowledge and unlearns them during the integration of new knowledge, thereby reducing knowledge conflicts within the LLM. By performing both steps simultaneously, CoME minimizes unintended knowledge transformations. This process is analogous to how humans enhance cognitive clarity by discarding irrelevant or erroneous memories.
Importantly, CoME achieves this without unnecessary loss of linguistic understanding, as we carefully preserve critical language-processing features shared between outdated and new knowledge. Furthermore, we limit the parameter space subject to modification during the unlearning process to minimize unnecessary parameter adjustments.

We apply CoME to state-of-art model editing methods, including MEMIT~\cite{meng2023massediting} and PMET~\cite{li2024pmet}, which are designed to mitigate overfitting and memory overhead issues in knowledge editing. We conduct large-scale knowledge editing experiments on 10,000 samples from the Counterfact~\cite{meng2023locating} and ZsRE~\cite{levy2017zero} datasets, utilizing the GPT-J (6B)~\cite{gpt-j} and LLaMA-3 (8B)~\cite{dubey2024llama3herdmodels}. The results demonstrate that applying CoME significantly improves the accuracy of knowledge updates. In particular, we show that CoME suppresses interference from outdated knowledge during inference, resulting in consistent and accurate responses, while maintaining the LLM's pre-existing capabilities.

Our main contributions are as follows:
\begin{itemize}
    \item We propose a new framework to mitigate conflicts between outdated and new knowledge in LLMs' knowledge editing. 
    \item We introduce unlearning to remove outdated knowledge while integrating new information, and we design an algorithm that applies unlearning selectively to relevant parameters. Our method is designed to complement and enhance existing model editing methods.
    \item Our experiments demonstrate that CoME suppresses interference from outdated knowledge, yielding more reliable and consistent responses. This highlights the framework’s ability to enhance the robustness of LLMs when handling updated information.
\end{itemize}

\section{Related Work}
\subsection{Knowledge Editing}
Existing knowledge editing methods can generally be divided into two categories: preserve and modify parameters.
One involves editing a model's knowledge without directly modifying its parameters. SERAC~\cite{mitchell2022memorybased} stores corrections in external memory and adjusts the model’s responses as needed. 
IKE~\cite{zheng2023edit} proposes a solution based on in-context learning, which enables knowledge modification without parameter updates. GRACE~\cite{hartvigsen2023aging} maps keys to the latent space of the model without changing the weights, constructing a local codebook for knowledge editing. While these methods are resilient to catastrophic forgetting due to the lack of parameter modifications, they require additional memory, which increases with the number of knowledge updates.

Early approaches that modify model parameters for knowledge editing often relied on fine-tuning techniques using multi-loss optimization, as proposed by \citet{sinitsin2020editable}. However, fine-tuning methods can lead to overfitting, prompting the development of hyperparameter-based optimization methods. Knowledge editor (KE)~\cite{decao2021editing} addresses this by utilizing a hypernetwork to edit specific knowledge without affecting unrelated knowledge. ROME~\cite{meng2023locating} identifies the multi-layer perceptrons (MLPs) where factual knowledge is stored and inserts new key-value pairs into those MLPs to modify the model’s knowledge. MEMIT~\cite{meng2023massediting} extends this approach, allowing the insertion of large volumes of knowledge simultaneously. PMET~\cite{li2024pmet} further optimizes the hidden states of both the multi-head self-attention (MHSA) and feed-forward network (FFN) layers to update the feed-forward weights efficiently.

\subsection{Unlearning}
The concept of machine unlearning, introduced by \citet{cao2015towards}, focuses on the removal of knowledge that has already been learned by a model. 
\citet{jang2022knowledge} employ gradient ascent to perform unlearning with the goal of alleviating privacy concerns, while \citet{eldan2023s} demonstrate unlearning by erasing specific knowledge related to the Harry Potter books from a model. \citet{chen2023unlearn} proposes freezing the LLM and introducing an unlearning layer to construct a forgotten model. \citet{yao-etal-2024-machine} presents a comprehensive framework for performing unlearning in LLMs using gradient ascent and KL divergence. \citet{hu2024separate} utilize parameter-efficient modules to preserve general model capabilities while removing untruthful or toxic information from LLMs.


\citet{ni2024forgettinglearningutilizingparametric} propose an approach that performs unlearning of existing knowledge before knowledge editing. However, this method is prone to overfitting due to its reliance on fine-tuning. In contrast, our approach is applied to state-of-the-art model editing methods that address such issues. By effectively removing outdated knowledge during the injection of new information, we mitigate conflicts between the two processes.

\section{Preliminaries}
\subsection{Model Editing}
The goal of model editing is to update the knowledge contained in LLM by replacing incorrect or outdated information with new knowledge. In this work, we focus on knowledge represented as triples consisting of a subject $s$, a relation $r$, and an object $o$. Our approach performs batch editing, where multiple pieces of knowledge are updated simultaneously. Specifically, given a model $f$ with parameters $\theta$, we update its parameters to $\theta^*$ by modifying $N$ pieces of knowledge in one step. The knowledge $G$ embedded in the model is represented as:
\begin{eqnarray}
G = \{(s_i, r_i, o_i), i \in [1,N]\}.
\end{eqnarray}

When editing knowledge, we replace the object in the outdated triple $(s, r, o)$ with a new object $o^*$, yielding updated knowledge $(s, r, o^*)$. The target knowledge $G^*$ that the updated model should encode is represented as:
\begin{eqnarray}
G^* = \{(s_i, r_i, o^*_i), i \in [1,N]\}.
\end{eqnarray}

For example, consider the case where $s_i = \texttt{``Motion,''}$ $r_i = \texttt{``manufactured by,''}$ and $o_i = \texttt{``Microsoft,''}$ which reflects an incorrect fact. The updated knowledge should modify the object to $o^*_i = \texttt{``Apple,''}$ while keeping the subject and relation intact. The prompt $x_i$ provided to the model might be ``\texttt{Motion, a product manufactured by},'' and the model's response should be updated to reflect the correct object $o^*_i$ rather than the incorrect $o_i$. Thus, the updated model must satisfy:
\begin{eqnarray}
f_{\theta^*}(x_i) = o^*_i, i \in [1,N].
\end{eqnarray}

If the model has been correctly edited, it satisfies \textbf{Efficacy}, a key attribute that should be prioritized in the editing process. Beyond efficacy, the following properties are essential for evaluating the quality of model editing:

\paragraph{Generality} ensures that the edited knowledge remains intact even when the prompt is paraphrased. This is measured by providing a paraphrased prompt $x^{gen}_i$ and checking whether the model still outputs the updated object $o^*_i$. For instance, if the paraphrased prompt is $x^{gen}_i = \texttt{``He}$ \texttt{was re-elected on the Hapoel HaMizrachi list in 1951. Motion, created by,''} the model should respond with $o^*_i$ to demonstrate that it retains the updated knowledge $(s_i, r_i, o^*_i)$ and applies it consistently across different prompts.

\paragraph{Locality} ensures that editing does not negatively impact unedited knowledge. The updated model must accurately modify only the target knowledge, leaving other unrelated information unchanged. For example, given a prompt that includes unchanged knowledge $x^{loc}_i =$ ``\texttt{Windows was developed by},'' the model’s response should remain consistent with the unedited knowledge $o_i$. This requirement can be formalized as:
\begin{eqnarray}
f_{\theta^*}(x^{loc}_i) = f_\theta(x^{loc}_i), i \in [1,N],
\end{eqnarray}
which ensures that the model’s responses to prompts involving unedited knowledge remain identical before and after the editing process.

\subsection{Locate-then-Edit}
Following the approach of \citet{meng2023massediting}, our goal is to efficiently update the weights of specific layers within the model in response to editing requests. Each edit request involves optimizing target vectors, which gradually adjust the weights of the layers.

We compute the update for one layer and then distribute it uniformly across the target layers. This allows us to update multiple layers efficiently with minimal computational overhead. Specifically, we focus on the final target layer $l$ among the set of target layers $T$. Given an input $x_i$, we calculate a replacement vector $z_i$ for the hidden state $h_i^l$ in layer $l$ as follows: $z_i = h_i^l + \delta_i$. The residual vector $\delta_i$, used to update $z_i$, is optimized via gradient descent:
\begin{equation} \label{eqn:opt_new}
\arg\min_{\delta_i} \frac{1}{P} \sum_{j=1}^{P} -\log \mathbb{P}_{f_\theta(h_i^l += \delta_i)}\left[ o_i^* \mid p_j + x_i \right],
\end{equation}
where $p_j$ represents the $P$ additional prompts introduced to enhance the diversity of inputs.

The computed update is then distributed across the target layers by modifying the MLP weights. Let $W$ represent the original weights, and $\hat{W}$ the updated weights. The incremental update $\Delta$ is added to the original weights, resulting in $\hat{W} = W + \Delta$. The incremental update $\Delta$ is calculated as follows:
\begin{eqnarray}
\Delta = R\hat{K}^T(C + \hat{K}\hat{K}^T)^{-1},
\end{eqnarray}
where $\hat{K}$ encodes the key associated with the target knowledge to be updated. The matrix $C \triangleq KK^T$ represents a set of previously memorized keys obtained through sampling. $R \triangleq \hat{V} - W\hat{K}$ represents the difference between the model's original knowledge representation $W\hat{K}$ and the target knowledge representation $\hat{V}$. This represents a set of values where the residual vector is distributed across the target layers using $\frac{\delta_i}{l - t + 1}, t \in T$.

\section{CoME: Conflict-free Model Editing}

As shown in Figure~\ref{fig:main}, we propose CoME that improves the accuracy of knowledge editing by utilizing parameter subtraction-based unlearning. Our method introduces three key steps to enhance existing locate-then-edit methods: (1) extracting parameters associated with outdated knowledge, (2) performing targeted unlearning during the integration of new knowledge, and (3) restricting the unlearning process to a specific parameter range to ensure that only essential portions are affected.

\subsection{Extraction of Outdated Knowledge Parameters}

To minimize conflicts between outdated knowledge and new knowledge, we remove the outdated information from the updated parameters $z_i$ before distributing the updates across the target layers. First, we obtain the parameters $\delta'_i$ that update the model with outdated knowledge in order to extract the parameters associated with this knowledge. This process closely mirrors the procedure for obtaining the parameters $\delta_i$ corresponding to the new knowledge. $\delta'_i$ is obtained by replacing the new knowledge $o^*$ with the outdated knowledge $o$ in Equation~\ref{eqn:opt_new} and learning accordingly. Therefore, it represents the parameters associated with outdated knowledge.
Inspired by \citet{ilharco2023editingmodelstaskarithmetic, zhang2023composing}, we hypothesize that subtracting the parameters associated with outdated knowledge from the model can facilitate effective unlearning of that knowledge.
By performing $z_i - \delta'_i$, we aim to remove the portions of the parameters updated with new knowledge that still contain outdated information. 


Following the insights from \citet{hu2024separate}, we assume that $\delta'_i$ not only encapsulates outdated knowledge but also encompasses the model's linguistic abilities. As shown in Equation \ref{eqn:opt_new}, both the outdated knowledge update vector $\delta'_i$ and the new knowledge update vector $\delta_i$ are trained using the same input prompt $x_i$. Therefore, both vectors inherently include the linguistic capacity necessary for the model to generate correct responses based on this information. If this shared capability is removed, the model’s ability to provide accurate responses to inputs may be compromised, making it essential to preserve this feature. To achieve this, we extract the common linguistic features by fusing $\delta_i$ and $\delta'_i$. Since the two linearly independent vectors span distinct hyperplanes, we obtain the direction vector $\vec{\delta}_i$ representing the common component by adding their normalized values:
\begin{eqnarray}
\vec{\delta}_i = \frac{\delta_i}{|\delta_i|} + \frac{\delta'_i}{|\delta'_i|}.
\end{eqnarray}

The common part of the outdated and new knowledge vectors is then extracted using vector projection:
\begin{eqnarray}
\delta''_i = \delta'_i \cdot \frac{\vec{\delta}_i}{|\vec{\delta}_i|}.
\end{eqnarray}

\subsection{Unlearning During Knowledge Update}

After extracting the common component, we subtract it from the outdated knowledge update vector. The remaining component, which encodes only outdated knowledge, is subtracted from the updated parameters:
\begin{eqnarray}
z'_i = z_i - \alpha(\delta'_i-\delta''_i),
\end{eqnarray}
where $\alpha$ is a hyperparameter controlling the weight of the subtraction operation\footnote{The process of determining the optimal $\alpha$ is detailed in Section~\ref{sec:alpha}.}.

\subsection{Restricting Unlearning to Critical Parameters}

Through the experiment shown in Figure~\ref{fig:alpha}, we confirm that unlearning outdated knowledge negatively affects Locality. To address this, inspired by \citet{gu2024modeleditingharmsgeneral}, we limit the scope of unlearning to only the parameters most influenced by outdated knowledge, leaving other knowledge unaffected. Specifically, we restrict unlearning to the top-p\% of parameters based on the magnitude of the unlearning update\footnote{The hyperparameter p is empirically set to 20 in this work.}. Parameters in the top-p\% are considered essential for unlearning, while the remaining parameters are treated as irrelevant. The final update for parameter $z'_i$ is as follows:
\begin{eqnarray}
z'_i =
\begin{cases} 
      z'_i & \text{if } (|\delta'_i-\delta''_i|) \text{ in the top-p\%}, \\
      z_i & \text{otherwise.}
\end{cases}
\end{eqnarray}

\begin{table*}[]
\centering
\renewcommand{\arraystretch}{1.4}
{\footnotesize
\begin{tabular}{lcccccc}
\toprule
\textbf{Method}                 & \textbf{Score}         & \textbf{Efficacy}            & \textbf{Generality}          & \textbf{Locality}         & \textbf{Fluency}              & \textbf{Consistency}         \\ \midrule
\textbf{GPT-J}   & 22.4          & 15.2 (0.7)          & 17.7 (0.6)          & 83.5 (0.5)          & 622.4 (0.3)          & 29.4 (0.2)          \\ \cdashlinelr{0-6}\noalign{\vskip 0.2ex}
FT-W             & 67.6          & 99.4 (0.1)          & 77.0 (0.7)            & 46.9 (0.6)          & 293.9 (2.4)          & 15.9 (0.3)          \\
FT & 35.6 & 29.0 (0.5) & 28.1 (0.4) & 71.4 (0.3) & 516.9 (0.7) & 10.4 (0.1)         \\
F-Learning & 38.1 & 30.5 (0.5) & 30.8 (0.4) & 73.7 (0.3) & 532.8 (0.7) & 12.8 (0.1)         \\
MEND             & 23.1          & 15.7 (0.7)          & 18.5 (0.7)          & \textbf{83.0 (0.5)}          & 618.4 (0.3)          & 31.1 (0.2)          \\
ROME             & 50.3          & 50.2 (1.0)          & 50.4 (0.8)          & 50.2 (0.6)          & 589.6 (0.5)          & 3.3 (0.0)           \\
MEMIT            & 85.8          & 98.9 (0.2)          & 88.6 (0.5)          & 73.7 (0.5) & 619.9 (0.3)          & 40.1 (0.2)          \\
PMET             & 86.2          & 99.5 (0.1)          & 92.8 (0.4)          & 71.4 (0.5)          & \textbf{620.0 (0.3)} & 40.6 (0.2)          \\ \hline
\textbf{$\text{CoME}_{\text{MEMIT}}$}       & \textbf{86.4} & 99.4 (0.1)          & 91.1 (0.2)          & 73.2 (0.3)          & 619.8 (0.1)          & \textbf{40.7 (0.1)} \\
\textbf{$\text{CoME}_{\text{PMET}}$}        & \textbf{86.4} & \textbf{99.8 (0.0)} & \textbf{95.3 (0.2)} & 70.3 (0.3)          & 618.9 (0.2)          & 40.3 (0.1)          \\
 \toprule \bottomrule
\textbf{LLaMA-3} & 15.0          & 9.6 (0.3)           & 11.8 (0.3)          & 87.6 (0.2)          & 628.4 (0.1)          & 26.3 (0.1)          \\ \cdashlinelr{0-6}\noalign{\vskip 0.2ex}
FT-W             & 42.9          & 37.5 (0.5)          & 36.6 (0.4)          & 62.8 (0.4)          & 437.5 (0.1)          & 4.7 (0.1)           \\
FT & 28.7 & 20.0 (0.4) & 23.4 (0.3) & 78.9 (0.3) & 613.9 (0.2) & 23.4 (0.1)         \\
F-Learning & 32.1 & 25.0 (0.1) & 23.9 (0.4) & \textbf{84.8 (0.2)} & 611.3 (0.3) & 23.6 (0.1)         \\
ROME             & 49.0          & 47.6 (0.5)          & 47.4 (0.5)          & 52.4 (0.5)         & 602.3 (0.0)          & 0.7 (0.0)           \\
MEMIT            & 78.2          & 94.9 (0.2)          & 90.5 (0.2)          & 59.6 (0.3)          & 608.8 (0.2)          & \textbf{42.5 (0.1)} \\
PMET             & 81.1          & 90.5 (0.3)          & 79.2 (0.3)          & 75.3 (0.3)          & 626.0 (0.1)          & 35.4 (0.1)          \\ \hline
\textbf{$\text{CoME}_{\text{MEMIT}}$}       & 78.2          & \textbf{95.7 (0.2)} & \textbf{91.3 (0.2)} & 59.0 (0.3)          & 610.9 (0.3)          & 41.0 (0.1)          \\
\textbf{$\text{CoME}_{\text{PMET}}$}        & \textbf{82.3} & 92.4 (0.3)          & 83.6 (0.3)          & 73.3 (0.3) & \textbf{627.9 (0.1)} & 36.8 (0.1)    \\ \bottomrule
\end{tabular}
}
\caption{10,000 Counterfact edits on GPT-J and LLaMA-3. The 95\% confidence interval is provided within parentheses.}
\label{tab:main_coun}
\end{table*}

\begin{table}[hbt!]
\centering
\renewcommand{\arraystretch}{1.4}
{\footnotesize
\begin{tabular}{lccc}
\toprule
\textbf{Method}                   & \textbf{Efficacy}         & \textbf{Generality}          & \textbf{Locality}            \\ \midrule
\textbf{GPT-J}     & 26.4 (0.6)          & 25.8 (0.5)          & 27.0 (0.5)          \\ \cdashlinelr{0-3}\noalign{\vskip 0.2ex}
FT-W               & 69.6 (0.6)          & 64.8 (0.6)          & 24.1 (0.5)          \\
FT               & 52.2 (0.4) & 49.6 (0.4) & 24.5 (0.2)          \\
F-Learning               & 58.8 (0.4) & 55.4 (0.4) & 24.8 (0.2)          \\
MEND               & 19.4 (0.5)          & 18.6 (0.5)          & 22.4 (0.5)          \\
ROME               & 21.0 (0.7)          & 19.6 (0.7)          & 0.9 (0.1)           \\
MEMIT              & 96.7 (0.3)          & 89.7 (0.5)          & \textbf{26.6 (0.5)}         \\
PMET               & 86.5 (0.3)          & 79.5 (0.3)          & 26.1 (0.3) \\ \hline
\textbf{$\text{CoME}_{\text{MEMIT}}$}         & \textbf{97.3 (0.1)} & \textbf{93.0 (0.2)} & 25.9 (0.2)          \\
\textbf{$\text{CoME}_{\text{PMET}}$}          & 89.4 (0.2)          & 83.1 (0.3)          & 26.3 (0.3) \\ \toprule \bottomrule
\textbf{LLaMA-3} & 40.9 (0.3)          & 36.3 (0.3)          & 37.6 (0.3)          \\ \cdashlinelr{0-3}\noalign{\vskip 0.2ex}
FT-W               & 19.3 (0.2)          & 17.5 (0.2)          & 9.9 (0.2)           \\ 
FT               & 61.5 (0.3) & 60.1 (0.3) & 44.2 (0.3)          \\
F-Learning               & 64.1 (0.4) & 61.5 (0.4) & 45.0 (0.3)          \\
ROME               & 0.0 (0.0)           & 0.0 (0.0)           & 0.1 (0.0)           \\
MEMIT              & 65.5 (0.3)          & 63.2 (0.3)          & 15.8 (0.2)          \\
PMET               & 90.2 (0.2)          & 87.4 (0.2)          & 47.0 (0.3)          \\ \hline
\textbf{$\text{CoME}_{\text{MEMIT}}$}         & 64.5 (0.3)          & 62.5 (0.4)          & 18.6 (0.2)          \\
\textbf{$\text{CoME}_{\text{PMET}}$}          & \textbf{90.6 (0.2)} & \textbf{87.8 (0.2)} & \textbf{47.4 (0.3)}
\\ \bottomrule
\end{tabular}
}
\caption{10,000 ZsRE edits on GPT-J and LLaMA-3.}
\label{tab:main_zsre}
\end{table}

\section{Experiments}
\subsection{Setup}
\paragraph{Datasets}
We adopt two widely used evaluation datasets from existing model editing research: Counterfact~\cite{meng2023locating} and ZsRE~\cite{levy2017zero}. The Counterfact dataset contains counterfactual knowledge, statements that have a lower generation probability than factual knowledge, which are provided as new knowledge for editing. To assess large-scale knowledge editing capabilities, we conduct experiments on 10,000 samples. ZsRE is a context-free question-answering dataset designed for zero-shot relation extraction. We extract 10,000 samples from ZsRE to evaluate the models' ability to accurately edit knowledge.

\paragraph{Metrics}
In the Counterfact dataset, we evaluate the models on Efficacy, Generality, and Locality, using success rates as metrics. Additionally, we assess the models' generative capabilities through Fluency and Consistency. Score is the harmonic mean of Efficacy, Generality, and Locality. Since ZsRE does not measure generative capabilities, we evaluate the models based only on accuracy in terms of Efficacy, Generality, and Locality. A detailed description of the evaluation metrics can be found in Appendix~\ref{sec:metric}.

\paragraph{Baselines}
To enable a direct comparison with existing model editing methods, we follow the baselines outlined in \citet{li2024pmet}. The first baseline is the unedited model. FT-W~\cite{zhu2020modifyingmemoriestransformermodels}, involves fine-tuning using weight decay for knowledge editing. FT fine-tunes all parameters of the base model. F-Learning~\cite{ni2024forgettinglearningutilizingparametric} is a fine-tuning-based approach that forgets existing knowledge and learns new knowledge. MEND~\cite{mitchell2022fastmodeleditingscale} leverages additional training data to fine-tune the model through a hypernetwork-based approach. ROME~\cite{meng2023locating} is an optimization-based method for single-editing tasks, while MEMIT~\cite{meng2023massediting} extends ROME to enable large-scale knowledge editing in a single pass. PMET~\cite{li2024pmet} optimizes both MHSA and FFN components simultaneously for knowledge editing.

\paragraph{Implementation Details}
We conduct our experiments using GPT-J (6B)~\cite{gpt-j} and LLaMA-3 (8B)~\cite{dubey2024llama3herdmodels}. The model checkpoints used are `EleutherAI/gpt-j-6B' and `meta-llama/LLaMA-3.1-8B', both of which are available on HuggingFace\footnote{\url{https://huggingface.co/}}. For GPT-J, following \citet{meng2023massediting}, we update layers \{3, 4, 5, 6, 7, 8\}. For LLaMA-3, following \citet{wang2023easyedit}, we update layers \{4, 5, 6, 7, 8\}. To estimate the covariance matrix $C$, we sample 10K times from WikiText in fp32 precision. For MEMIT, we set the covariance adjustment factor $\lambda = 15000$, and for PMET, we set $\lambda = 6000$. All experiments are performed using a single RTX A6000 GPU. GPT-J is run in fp32, while LLaMA-3 uses fp16 due to memory constraints. 
Unlike MEMIT, in the PMET setting, only the weights of the FFN are updated separately, so CoME is applied to the FFN residual vector $\delta^{FFN}_i$.
Further implementation details can be found in the official MEMIT\footnote{\url{https://github.com/kmeng01/memit}} and PMET\footnote{\url{https://github.com/xpq-tech/PMET}} repositories.

\begin{figure*}[hbt!]
\centering 
\includegraphics[width=0.97\linewidth]{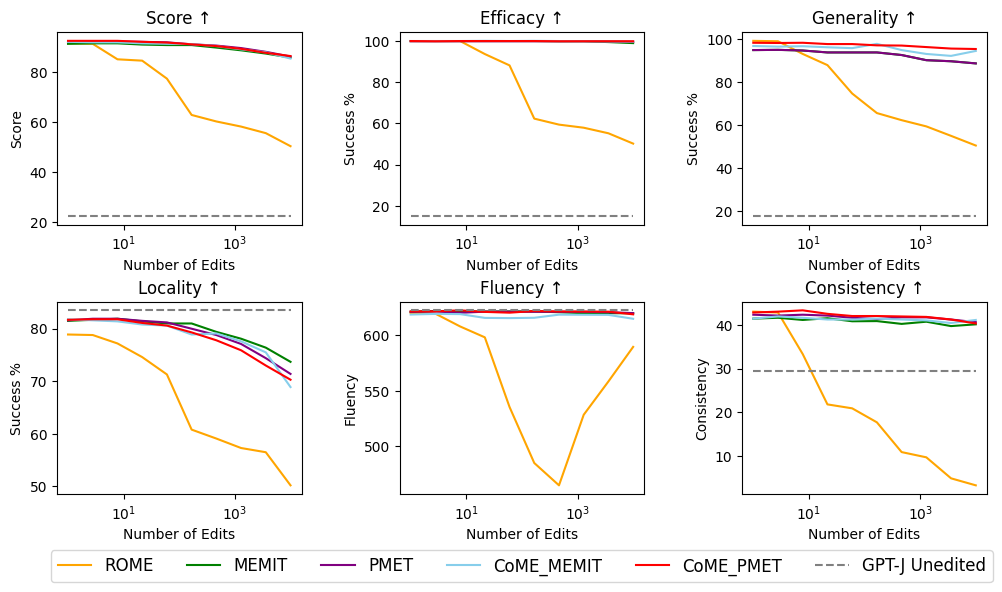}
\caption{Scaling curves that represent editing performance based on the size of edits. These experiments are conducted on the Counterfact dataset using GPT-J.}
\label{fig:number} 
\end{figure*}

\subsection{Main Results}

\paragraph{Editing Knowledge in Counterfact}
Table~\ref{tab:main_coun} presents the editing performance of CoME on 10,000 samples from the Counterfact dataset. Both $\text{CoME}_{\text{MEMIT}}$ and $\text{CoME}_{\text{PMET}}$ improve Score, which evaluates the overall performance of editing. On GPT-J, both methods achieve Score of 86.4, compared to 85.8 for MEMIT and 86.2 for PMET, demonstrating the efficacy of our approach. Similarly, on LLaMA-3, $\text{CoME}_{\text{PMET}}$ achieves 82.3, outperforming PMET of 81.1. These results show that by removing outdated knowledge, our method enhances the model’s ability to handle new knowledge. The most notable improvement arises in the accuracy of newly updated knowledge, particularly in terms of Efficacy and Generality. Not only does the accuracy of the edited knowledge increase, but interference from outdated knowledge is minimized, resulting in higher overall performance.

In contrast, Locality, which measures the preservation of unrelated knowledge, slightly decreases compared to MEMIT and PMET. This trade-off between editing accuracy and Locality is expected, as our primary objective is to inject new knowledge rather than minimize changes to the model. Furthermore, Fluency and Consistency of the model’s outputs are maintained at levels comparable to the original model, further supporting the robustness of our method. Appendix~\ref{sec:case} presents a case study demonstrating how CoME enhances the utilization of new knowledge by unlearning outdated knowledge.

\paragraph{Editing Knowledge in ZsRE}
Table~\ref{tab:main_zsre} shows the performance of our method on 10,000 ZsRE samples using GPT-J and LLaMA-3. Similar to the results on the Counterfact dataset, $\text{CoME}_{\text{MEMIT}}$ and $\text{CoME}_{\text{PMET}}$ demonstrate superior performance in Efficacy and Generality on both models. For GPT-J, $\text{CoME}_{\text{PMET}}$ achieves Efficacy of 89.4 and Generality of 83.1, both surpassing the results of baseline PMET. These outcomes suggest that our method effectively integrates new knowledge while minimizing the influence of outdated information. 

In terms of Locality, the results on ZsRE show significant improvements compared to the Counterfact dataset. $\text{CoME}_{\text{PMET}}$ achieves the highest Locality scores on both models, indicating that our approach reduces the negative impact on unrelated knowledge. Particularly on LLaMA-3, $\text{CoME}_{\text{PMET}}$ not only updates knowledge but also improves the model’s ability to generate factual responses compared to the original model.

\subsection{Analysis}

\begin{table*}[hbt!]
\centering
\renewcommand{\arraystretch}{1.4}
{\footnotesize
\begin{tabular}{lcccccc}
\toprule
\textbf{Method}                        & \textbf{Score} & \textbf{Efficacy} & \textbf{Generality} & \textbf{Locality} & \textbf{Fluency} & \textbf{Consistency} \\ \midrule
GPT-J          & 22.4           & 15.2 (0.7)        & 17.7 (0.6)          & 83.5 (0.5)           & 622.4 (0.3)      & 29.4 (0.2)           \\ \hline
\textbf{$\text{CoME}_{\text{MEMIT}}$}              & 86.4           & 99.4 (0.1)        & 91.1 (0.2)          & 73.2 (0.3)           & 619.8 (0.1)      & 40.7 (0.1)           \\ \midrule
w/o $\delta'$   & 85.7 ↓         & 99.1 (0.1) ↓      & 88.6 (0.3) ↓        & 73.5 (0.3) ↑         & 619.3 (0.2) ↓    & 40.0 (0.1) ↓         \\
w/o $\delta''$ & 84.1 ↓         & 99.1 (0.1) ↓      & 89.3 (0.3) ↓        & 69.6 (0.3) ↓         & 621.1 (0.1) ↑    & 40.2 (0.1) ↓         \\ 
w/o restriction        & 85.4 ↓         & 99.6 (0.1) ↑      & 94.4 (0.2) ↑        & 68.9 (0.3) ↓         & 614.9 (0.2) ↓    & 41.1 (0.1) ↑         \\ \midrule
\textbf{$\text{CoME}_{\text{PMET}}$}               & 86.4           & 99.8 (0.0)        & 95.3 (0.2)          & 70.3 (0.3)           & 618.9 (0.2)      & 40.3 (0.1)           \\ \hline
w/o $\delta'$    & 84.1 ↓         & 99.5 (0.1) ↓      & 96.6 (0.1) ↑        & 65.5 (0.3) ↓         & 619.8 (0.2) ↑    & 42.5 (0.1) ↑         \\
w/o $\delta''$ & 85.4 ↓         & 99.5 (0.1) ↓      & 93.4 (0.2) ↓        & 69.7 (0.3) ↓         & 619.8 (0.2) ↑    & 41.4 (0.1) ↑         \\ 
w/o restriction        & 85.9 ↓         & 99.7 (0.1) ↓      & 94.6 (0.2) ↓        & 69.8 (0.3) ↓         & 619.3 (0.3) ↑    & 41.0 (0.1) ↑        
 \\ \bottomrule
\end{tabular}
}
\caption{The results of ablation study. $\delta'$ represents the parameters that update outdated knowledge, while $\delta''$ corresponds to the shared linguistic capabilities. Restriction limits the parameter space and subjects it to unlearning. The experiments are conducted using GPT-J on 10,000 Counterfact samples. }
\label{tab:ablation}
\end{table*}

\paragraph{Number of Edits}

Figure~\ref{fig:number} illustrates the performance of the model as the number of simultaneous edits increases. The results show that $\text{CoME}_{\text{MEMIT}}$ and $\text{CoME}_{\text{PMET}}$ remain robust in terms of Efficacy and Generality, even as the number of edits increases. Our method ensures a high success rate for Generality, even when the number of edits is low, and maintains initial performance levels as the number of edits grows. However, as with other methods, Locality begins to decline sharply once the number of edits exceeds a certain threshold. In terms of Fluency and Consistency, our methods perform similarly to or exceed the original model’s performance, unlike ROME, which experiences significant drops in language generation quality as the number of edits increases.

\paragraph{Ablation Study}
The results of the ablation study, presented in Table~\ref{tab:ablation}, examine the effects of unlearning and the application of restricting unlearning parameters on $\text{CoME}_{\text{MEMIT}}$ and $\text{CoME}_{\text{PMET}}$. We analyze the impact of removing each component: $\delta'$, $\delta''$, and restricting unlearning parameters.


Excluding $\delta'$, we observe a decline in performance across most metrics, particularly in Generality and Efficacy. Notably, in $\text{CoME}_{\text{MEMIT}}$, performance drops significantly from 91.1 to 88.6. This suggests that the removal of outdated knowledge plays a crucial role in improving the accuracy of knowledge editing.

Excluding $\delta''$ primarily affects Locality, where we observe significant performance degradation. This suggests that $\delta''$ plays a vital role in preserving the model’s ability to handle unrelated information. On the other hand, Fluency shows an upward trend, likely due to the increased capacity to handle structured knowledge, which comes at the cost of penalties in generation fluency.

Excluding restricting the unlearning parameter method leads to the greatest drop in Locality, while Efficacy and Generality are only slightly affected. This shows that unlearning is effectively performed only on the top-p\% of parameters where outdated knowledge resides, preventing unnecessary parameter updates without sacrificing accuracy.

\paragraph{Unlearning Weight Variation} \label{sec:alpha}

\begin{figure}[]
\centering 
\includegraphics[width=0.99\columnwidth]{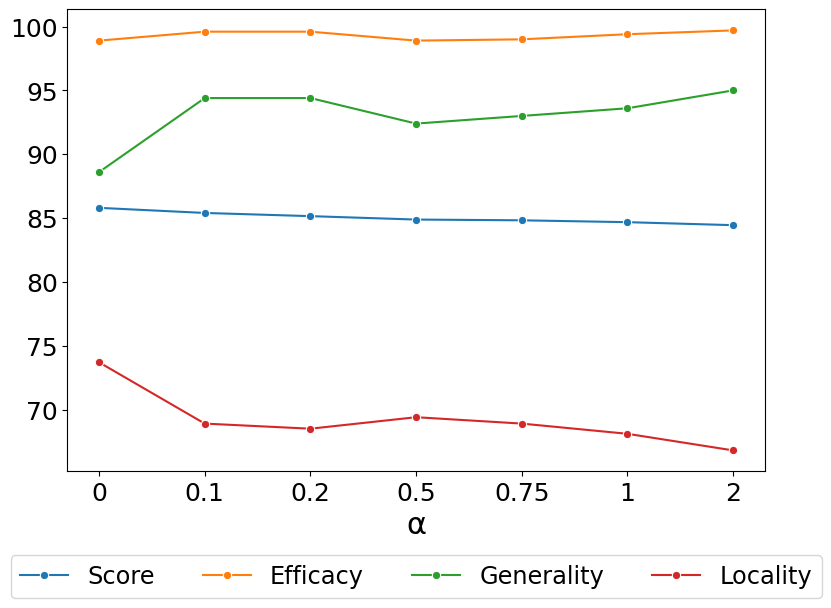}
\caption{Effect of unlearning weight variation in $\text{CoME}_{\text{MEMIT}}$. The experiments are conducted using GPT-J and 10,000 Counterfact samples.}
\label{fig:alpha} 
\end{figure}

To control the degree of outdated knowledge removal, we introduce the hyperparameter $\alpha$. Figure~\ref{fig:alpha} shows the effect of varying $\alpha$ from 0 to 2 on performance metrics such as Score, Efficacy, Generality, and Locality. We observe that both Efficacy and Generality increase as $\alpha$ rises, indicating that more effective removal of outdated knowledge improves model performance. However, Locality decreases as $\alpha$ increases, suggesting that excessive knowledge removal may negatively impact unrelated information. Based on these findings, we use $\alpha = 0.1$ as the default setting, and restrict the unlearning scope to minimize the drop in Locality.

\section{Conclusion}

In this paper, we proposed CoME to address the conflict between outdated and new knowledge that can arise during the editing process in LLMs. CoME enhanced the accuracy of knowledge editing by simultaneously unlearning outdated knowledge and integrating new information. Experiments showed that our method improved the editing accuracy of existing model editing methods and successfully integrated new knowledge. This approach can be an effective solution for correcting inaccurate or biased information in large language models, and we expect it to make significant contributions to improving the reliability and consistency of LLMs.

\section*{Limitations}

While CoME successfully enhances the usability of new knowledge by removing outdated information, several limitations must be acknowledged:

\begin{itemize}
    \item The unlearning process requires additional computational resources. Since CoME introduces a separate stage to remove outdated knowledge, it incurs higher computational costs than traditional model editing techniques.
    \item CoME is designed to remove outdated or false knowledge, which may not always be desirable in cases of temporal knowledge. For example, older information that reflects past realities can still be useful in certain contexts. 
\end{itemize}



\section*{Ethical Considerations}

Our research aims to enhance the reliability and safety of LLMs by addressing issues stemming from the retention of incorrect or biased information. By developing and improving model editing methods, we seek to contribute to the responsible use of LLMs, particularly in mitigating the spread of misinformation and harmful biases. However, it is essential to recognize that any modification to a model’s knowledge must be handled with caution, ensuring that only erroneous or biased information is removed while preserving the integrity of factual content. Ensuring that model editing is performed transparently and based on clearly defined ethical guidelines will be critical as this technology develops.

\section*{Acknowledgments}

This work was supported by Institute for Information \& communications Technology Promotion(IITP) grant funded by the Korea government(MSIT) (RS-2024-00398115, Research on the reliability and coherence of outcomes produced by Generative AI). This research was supported by Basic Science Research Program through the National Research Foundation of Korea(NRF) funded by the Ministry of Education(NRF-2021R1A6A1A03045425). This work was supported by ICT Creative Consilience Program through the Institute of Information \& Communications Technology Planning \& Evaluation(IITP) grant funded by the Korea government(MSIT)(IITP-2025-RS-2020-II201819).


\bibliography{main}

\begin{thebibliography}{49}
\providecommand{\natexlab}[1]{#1}

\bibitem[{Alves and Bueno(2017)}]{alves2017retroactive}
Marcus Vinicius~Costa Alves and Orlando Francisco~Amodeo Bueno. 2017.
\newblock Retroactive interference: forgetting as an interruption of memory consolidation.
\newblock \emph{Trends in Psychology}, 25:1043--1054.

\bibitem[{Bjork and Bjork(1996)}]{bjork1996continuing}
Elizabeth~Ligon Bjork and Robert~A Bjork. 1996.
\newblock Continuing influences of to-be-forgotten information.
\newblock \emph{Consciousness and cognition}, 5(1-2):176--196.

\bibitem[{Cao et~al.(2021{\natexlab{a}})Cao, Lin, Han, Sun, Yan, Liao, Xue, and Xu}]{cao2021knowledgeableeducatedguessrevisiting}
Boxi Cao, Hongyu Lin, Xianpei Han, Le~Sun, Lingyong Yan, Meng Liao, Tong Xue, and Jin Xu. 2021{\natexlab{a}}.
\newblock \href {https://arxiv.org/abs/2106.09231} {Knowledgeable or educated guess? revisiting language models as knowledge bases}.
\newblock \emph{Preprint}, arXiv:2106.09231.

\bibitem[{Cao et~al.(2021{\natexlab{b}})Cao, Aziz, and Titov}]{decao2021editing}
Nicola~De Cao, Wilker Aziz, and Ivan Titov. 2021{\natexlab{b}}.
\newblock \href {https://arxiv.org/abs/2104.08164} {Editing factual knowledge in language models}.
\newblock \emph{Preprint}, arXiv:2104.08164.

\bibitem[{Cao and Yang(2015)}]{cao2015towards}
Yinzhi Cao and Junfeng Yang. 2015.
\newblock Towards making systems forget with machine unlearning.
\newblock In \emph{2015 IEEE symposium on security and privacy}, pages 463--480. IEEE.

\bibitem[{Chen and Yang(2023)}]{chen2023unlearn}
Jiaao Chen and Diyi Yang. 2023.
\newblock Unlearn what you want to forget: Efficient unlearning for llms.
\newblock \emph{arXiv preprint arXiv:2310.20150}.

\bibitem[{Chen et~al.(2024)Chen, Li, Xiao, and Liu}]{chen2024largelanguagemodelbias}
Ruizhe Chen, Yichen Li, Zikai Xiao, and Zuozhu Liu. 2024.
\newblock \href {https://arxiv.org/abs/2405.09341} {Large language model bias mitigation from the perspective of knowledge editing}.
\newblock \emph{Preprint}, arXiv:2405.09341.

\bibitem[{Dai et~al.(2022)Dai, Dong, Hao, Sui, Chang, and Wei}]{dai2022knowledgeneuronspretrainedtransformers}
Damai Dai, Li~Dong, Yaru Hao, Zhifang Sui, Baobao Chang, and Furu Wei. 2022.
\newblock \href {https://arxiv.org/abs/2104.08696} {Knowledge neurons in pretrained transformers}.
\newblock \emph{Preprint}, arXiv:2104.08696.

\bibitem[{Eldan and Russinovich(2023)}]{eldan2023s}
Ronen Eldan and Mark Russinovich. 2023.
\newblock Who's harry potter? approximate unlearning in llms.
\newblock \emph{arXiv preprint arXiv:2310.02238}.

\bibitem[{Geiselman et~al.(1983)Geiselman, Bjork, and Fishman}]{geiselman1983disrupted}
Ralph~E Geiselman, Robert~A Bjork, and Deborah~L Fishman. 1983.
\newblock Disrupted retrieval in directed forgetting: a link with posthypnotic amnesia.
\newblock \emph{Journal of Experimental Psychology: General}, 112(1):58.

\bibitem[{Gu et~al.(2024)Gu, Xu, Ma, Lu, Ling, Chang, and Peng}]{gu2024modeleditingharmsgeneral}
Jia-Chen Gu, Hao-Xiang Xu, Jun-Yu Ma, Pan Lu, Zhen-Hua Ling, Kai-Wei Chang, and Nanyun Peng. 2024.
\newblock \href {https://arxiv.org/abs/2401.04700} {Model editing harms general abilities of large language models: Regularization to the rescue}.
\newblock \emph{Preprint}, arXiv:2401.04700.

\bibitem[{Hao et~al.(2021)Hao, Dong, Wei, and Xu}]{hao2021selfattention}
Yaru Hao, Li~Dong, Furu Wei, and Ke~Xu. 2021.
\newblock \href {https://arxiv.org/abs/2004.11207} {Self-attention attribution: Interpreting information interactions inside transformer}.
\newblock \emph{Preprint}, arXiv:2004.11207.

\bibitem[{Hartvigsen et~al.(2023)Hartvigsen, Sankaranarayanan, Palangi, Kim, and Ghassemi}]{hartvigsen2023aging}
Thomas Hartvigsen, Swami Sankaranarayanan, Hamid Palangi, Yoon Kim, and Marzyeh Ghassemi. 2023.
\newblock \href {https://arxiv.org/abs/2211.11031} {Aging with grace: Lifelong model editing with discrete key-value adaptors}.
\newblock \emph{Preprint}, arXiv:2211.11031.

\bibitem[{Hase et~al.(2021)Hase, Diab, Celikyilmaz, Li, Kozareva, Stoyanov, Bansal, and Iyer}]{hase2021language}
Peter Hase, Mona Diab, Asli Celikyilmaz, Xian Li, Zornitsa Kozareva, Veselin Stoyanov, Mohit Bansal, and Srinivasan Iyer. 2021.
\newblock \href {https://arxiv.org/abs/2111.13654} {Do language models have beliefs? methods for detecting, updating, and visualizing model beliefs}.
\newblock \emph{Preprint}, arXiv:2111.13654.

\bibitem[{Haviv et~al.(2023)Haviv, Cohen, Gidron, Schuster, Goldberg, and Geva}]{haviv2023understanding}
Adi Haviv, Ido Cohen, Jacob Gidron, Roei Schuster, Yoav Goldberg, and Mor Geva. 2023.
\newblock \href {https://arxiv.org/abs/2210.03588} {Understanding transformer memorization recall through idioms}.
\newblock \emph{Preprint}, arXiv:2210.03588.

\bibitem[{Hernandez et~al.(2023)Hernandez, Li, and Andreas}]{hernandez2023inspecting}
Evan Hernandez, Belinda~Z. Li, and Jacob Andreas. 2023.
\newblock \href {https://arxiv.org/abs/2304.00740} {Inspecting and editing knowledge representations in language models}.
\newblock \emph{Preprint}, arXiv:2304.00740.

\bibitem[{Hu et~al.(2024{\natexlab{a}})Hu, Cao, Chen, Liu, and Zhao}]{hu2024wilkewiselayerknowledgeeditor}
Chenhui Hu, Pengfei Cao, Yubo Chen, Kang Liu, and Jun Zhao. 2024{\natexlab{a}}.
\newblock \href {https://arxiv.org/abs/2402.10987} {Wilke: Wise-layer knowledge editor for lifelong knowledge editing}.
\newblock \emph{Preprint}, arXiv:2402.10987.

\bibitem[{Hu et~al.(2024{\natexlab{b}})Hu, Li, Hu, Zheng, Liu, and Zhang}]{hu2024separate}
Xinshuo Hu, Dongfang Li, Baotian Hu, Zihao Zheng, Zhenyu Liu, and Min Zhang. 2024{\natexlab{b}}.
\newblock Separate the wheat from the chaff: Model deficiency unlearning via parameter-efficient module operation.
\newblock In \emph{Proceedings of the AAAI Conference on Artificial Intelligence}, volume~38, pages 18252--18260.

\bibitem[{Ilharco et~al.(2023)Ilharco, Ribeiro, Wortsman, Gururangan, Schmidt, Hajishirzi, and Farhadi}]{ilharco2023editingmodelstaskarithmetic}
Gabriel Ilharco, Marco~Tulio Ribeiro, Mitchell Wortsman, Suchin Gururangan, Ludwig Schmidt, Hannaneh Hajishirzi, and Ali Farhadi. 2023.
\newblock \href {https://arxiv.org/abs/2212.04089} {Editing models with task arithmetic}.
\newblock \emph{Preprint}, arXiv:2212.04089.

\bibitem[{Jang et~al.(2022)Jang, Yoon, Yang, Cha, Lee, Logeswaran, and Seo}]{jang2022knowledge}
Joel Jang, Dongkeun Yoon, Sohee Yang, Sungmin Cha, Moontae Lee, Lajanugen Logeswaran, and Minjoon Seo. 2022.
\newblock Knowledge unlearning for mitigating privacy risks in language models.
\newblock \emph{arXiv preprint arXiv:2210.01504}.

\bibitem[{Ji et~al.(2023)Ji, Lee, Frieske, Yu, Su, Xu, Ishii, Bang, Madotto, and Fung}]{Ji_2023}
Ziwei Ji, Nayeon Lee, Rita Frieske, Tiezheng Yu, Dan Su, Yan Xu, Etsuko Ishii, Ye~Jin Bang, Andrea Madotto, and Pascale Fung. 2023.
\newblock \href {https://doi.org/10.1145/3571730} {Survey of hallucination in natural language generation}.
\newblock \emph{ACM Computing Surveys}, 55(12):1–38.

\bibitem[{Jiang et~al.(2023)Jiang, Sablayrolles, Mensch, Bamford, Chaplot, Casas, Bressand, Lengyel, Lample, Saulnier et~al.}]{jiang2023mistral}
Albert~Q Jiang, Alexandre Sablayrolles, Arthur Mensch, Chris Bamford, Devendra~Singh Chaplot, Diego de~las Casas, Florian Bressand, Gianna Lengyel, Guillaume Lample, Lucile Saulnier, et~al. 2023.
\newblock Mistral 7b.
\newblock \emph{arXiv preprint arXiv:2310.06825}.

\bibitem[{Kliegl and B{\"a}uml(2021)}]{kliegl2021mechanisms}
Oliver Kliegl and Karl-Heinz~T B{\"a}uml. 2021.
\newblock The mechanisms underlying interference and inhibition: A review of current behavioral and neuroimaging research.
\newblock \emph{Brain Sciences}, 11(9):1246.

\bibitem[{Levy et~al.(2017)Levy, Seo, Choi, and Zettlemoyer}]{levy2017zero}
Omer Levy, Minjoon Seo, Eunsol Choi, and Luke Zettlemoyer. 2017.
\newblock Zero-shot relation extraction via reading comprehension.
\newblock \emph{arXiv preprint arXiv:1706.04115}.

\bibitem[{Li et~al.(2024{\natexlab{a}})Li, Li, Song, Yang, Ma, and Yu}]{li2024pmet}
Xiaopeng Li, Shasha Li, Shezheng Song, Jing Yang, Jun Ma, and Jie Yu. 2024{\natexlab{a}}.
\newblock Pmet: Precise model editing in a transformer.
\newblock In \emph{Proceedings of the AAAI Conference on Artificial Intelligence}, volume~38, pages 18564--18572.

\bibitem[{Li et~al.(2024{\natexlab{b}})Li, Zhang, Yao, Wang, Chen, and Chen}]{li2024unveilingpitfallsknowledgeediting}
Zhoubo Li, Ningyu Zhang, Yunzhi Yao, Mengru Wang, Xi~Chen, and Huajun Chen. 2024{\natexlab{b}}.
\newblock \href {https://arxiv.org/abs/2310.02129} {Unveiling the pitfalls of knowledge editing for large language models}.
\newblock \emph{Preprint}, arXiv:2310.02129.

\bibitem[{{Llama Team}(2024)}]{dubey2024llama3herdmodels}
{Llama Team}. 2024.
\newblock \href {https://arxiv.org/abs/2407.21783} {The llama 3 herd of models}.
\newblock \emph{Preprint}, arXiv:2407.21783.

\bibitem[{Meng et~al.(2023{\natexlab{a}})Meng, Bau, Andonian, and Belinkov}]{meng2023locating}
Kevin Meng, David Bau, Alex Andonian, and Yonatan Belinkov. 2023{\natexlab{a}}.
\newblock \href {https://arxiv.org/abs/2202.05262} {Locating and editing factual associations in gpt}.
\newblock \emph{Preprint}, arXiv:2202.05262.

\bibitem[{Meng et~al.(2023{\natexlab{b}})Meng, Sharma, Andonian, Belinkov, and Bau}]{meng2023massediting}
Kevin Meng, Arnab~Sen Sharma, Alex Andonian, Yonatan Belinkov, and David Bau. 2023{\natexlab{b}}.
\newblock \href {https://arxiv.org/abs/2210.07229} {Mass-editing memory in a transformer}.
\newblock \emph{Preprint}, arXiv:2210.07229.

\bibitem[{Mitchell et~al.(2022{\natexlab{a}})Mitchell, Lin, Bosselut, Finn, and Manning}]{mitchell2022fastmodeleditingscale}
Eric Mitchell, Charles Lin, Antoine Bosselut, Chelsea Finn, and Christopher~D. Manning. 2022{\natexlab{a}}.
\newblock \href {https://arxiv.org/abs/2110.11309} {Fast model editing at scale}.
\newblock \emph{Preprint}, arXiv:2110.11309.

\bibitem[{Mitchell et~al.(2022{\natexlab{b}})Mitchell, Lin, Bosselut, Manning, and Finn}]{mitchell2022memorybased}
Eric Mitchell, Charles Lin, Antoine Bosselut, Christopher~D. Manning, and Chelsea Finn. 2022{\natexlab{b}}.
\newblock \href {https://arxiv.org/abs/2206.06520} {Memory-based model editing at scale}.
\newblock \emph{Preprint}, arXiv:2206.06520.

\bibitem[{Mousavi et~al.(2024)Mousavi, Alghisi, and Riccardi}]{mousavi2024llm}
Seyed~Mahed Mousavi, Simone Alghisi, and Giuseppe Riccardi. 2024.
\newblock \href {https://arxiv.org/abs/2404.08700} {Is your llm outdated? benchmarking llms \& alignment algorithms for time-sensitive knowledge}.
\newblock \emph{Preprint}, arXiv:2404.08700.

\bibitem[{Ni et~al.(2024)Ni, Chen, Li, Hu, Xu, and Yang}]{ni2024forgettinglearningutilizingparametric}
Shiwen Ni, Dingwei Chen, Chengming Li, Xiping Hu, Ruifeng Xu, and Min Yang. 2024.
\newblock \href {https://arxiv.org/abs/2311.08011} {Forgetting before learning: Utilizing parametric arithmetic for knowledge updating in large language models}.
\newblock \emph{Preprint}, arXiv:2311.08011.

\bibitem[{OpenAI(2023)}]{openai2023gpt4}
OpenAI. 2023.
\newblock \href {https://arxiv.org/abs/2303.08774} {Gpt-4 technical report}.
\newblock \emph{Preprint}, arXiv:2303.08774.

\bibitem[{Pagnoni et~al.(2021)Pagnoni, Balachandran, and Tsvetkov}]{pagnoni2021understanding}
Artidoro Pagnoni, Vidhisha Balachandran, and Yulia Tsvetkov. 2021.
\newblock \href {https://arxiv.org/abs/2104.13346} {Understanding factuality in abstractive summarization with frank: A benchmark for factuality metrics}.
\newblock \emph{Preprint}, arXiv:2104.13346.

\bibitem[{Pinter and Elhadad(2023)}]{pinter2023emptyingoceanspoonedit}
Yuval Pinter and Michael Elhadad. 2023.
\newblock \href {https://arxiv.org/abs/2310.11958} {Emptying the ocean with a spoon: Should we edit models?}
\newblock \emph{Preprint}, arXiv:2310.11958.

\bibitem[{Sharma et~al.(2024)Sharma, Atkinson, and Bau}]{sharma2024locatingeditingfactualassociations}
Arnab~Sen Sharma, David Atkinson, and David Bau. 2024.
\newblock \href {https://arxiv.org/abs/2404.03646} {Locating and editing factual associations in mamba}.
\newblock \emph{Preprint}, arXiv:2404.03646.

\bibitem[{Sinitsin et~al.(2020)Sinitsin, Plokhotnyuk, Pyrkin, Popov, and Babenko}]{sinitsin2020editable}
Anton Sinitsin, Vsevolod Plokhotnyuk, Dmitriy Pyrkin, Sergei Popov, and Artem Babenko. 2020.
\newblock Editable neural networks.
\newblock \emph{arXiv preprint arXiv:2004.00345}.

\bibitem[{Wang and Komatsuzaki(2021)}]{gpt-j}
Ben Wang and Aran Komatsuzaki. 2021.
\newblock {GPT-J-6B: A 6 Billion Parameter Autoregressive Language Model}.
\newblock \url{https://github.com/kingoflolz/mesh-transformer-jax}.

\bibitem[{Wang et~al.(2024)Wang, Zhang, Xu, Xi, Deng, Yao, Zhang, Yang, Wang, and Chen}]{wang2024detoxifyinglargelanguagemodels}
Mengru Wang, Ningyu Zhang, Ziwen Xu, Zekun Xi, Shumin Deng, Yunzhi Yao, Qishen Zhang, Linyi Yang, Jindong Wang, and Huajun Chen. 2024.
\newblock \href {https://arxiv.org/abs/2403.14472} {Detoxifying large language models via knowledge editing}.
\newblock \emph{Preprint}, arXiv:2403.14472.

\bibitem[{Wang et~al.(2023{\natexlab{a}})Wang, Zhang, Tian, Xi, Yao, Xu, Wang, Mao, Wang, Cheng et~al.}]{wang2023easyedit}
Peng Wang, Ningyu Zhang, Bozhong Tian, Zekun Xi, Yunzhi Yao, Ziwen Xu, Mengru Wang, Shengyu Mao, Xiaohan Wang, Siyuan Cheng, et~al. 2023{\natexlab{a}}.
\newblock Easyedit: An easy-to-use knowledge editing framework for large language models.
\newblock \emph{arXiv preprint arXiv:2308.07269}.

\bibitem[{Wang et~al.(2023{\natexlab{b}})Wang, Zhu, Liu, Zheng, Chen, and Li}]{wang2023knowledge}
Song Wang, Yaochen Zhu, Haochen Liu, Zaiyi Zheng, Chen Chen, and Jundong Li. 2023{\natexlab{b}}.
\newblock \href {https://arxiv.org/abs/2310.16218} {Knowledge editing for large language models: A survey}.
\newblock \emph{Preprint}, arXiv:2310.16218.

\bibitem[{Wixted(2004)}]{wixted2004psychology}
John~T Wixted. 2004.
\newblock The psychology and neuroscience of forgetting.
\newblock \emph{Annu. Rev. Psychol.}, 55(1):235--269.

\bibitem[{Yao et~al.(2024)Yao, Chien, Du, Niu, Wang, Cheng, and Yue}]{yao-etal-2024-machine}
Jin Yao, Eli Chien, Minxin Du, Xinyao Niu, Tianhao Wang, Zezhou Cheng, and Xiang Yue. 2024.
\newblock \href {https://doi.org/10.18653/v1/2024.acl-long.457} {Machine unlearning of pre-trained large language models}.
\newblock In \emph{Proceedings of the 62nd Annual Meeting of the Association for Computational Linguistics (Volume 1: Long Papers)}, pages 8403--8419, Bangkok, Thailand. Association for Computational Linguistics.

\bibitem[{Yao et~al.(2023)Yao, Wang, Tian, Cheng, Li, Deng, Chen, and Zhang}]{yao2023editinglargelanguagemodels}
Yunzhi Yao, Peng Wang, Bozhong Tian, Siyuan Cheng, Zhoubo Li, Shumin Deng, Huajun Chen, and Ningyu Zhang. 2023.
\newblock \href {https://arxiv.org/abs/2305.13172} {Editing large language models: Problems, methods, and opportunities}.
\newblock \emph{Preprint}, arXiv:2305.13172.

\bibitem[{Zhang et~al.(2023)Zhang, Liu, He et~al.}]{zhang2023composing}
Jinghan Zhang, Junteng Liu, Junxian He, et~al. 2023.
\newblock Composing parameter-efficient modules with arithmetic operation.
\newblock \emph{Advances in Neural Information Processing Systems}, 36:12589--12610.

\bibitem[{Zhang et~al.(2024)Zhang, Yao, Tian, Wang, Deng, Wang, Xi, Mao, Zhang, Ni, Cheng, Xu, Xu, Gu, Jiang, Xie, Huang, Liang, Zhang, Zhu, Zhou, and Chen}]{zhang2024comprehensive}
Ningyu Zhang, Yunzhi Yao, Bozhong Tian, Peng Wang, Shumin Deng, Mengru Wang, Zekun Xi, Shengyu Mao, Jintian Zhang, Yuansheng Ni, Siyuan Cheng, Ziwen Xu, Xin Xu, Jia-Chen Gu, Yong Jiang, Pengjun Xie, Fei Huang, Lei Liang, Zhiqiang Zhang, Xiaowei Zhu, Jun Zhou, and Huajun Chen. 2024.
\newblock \href {https://arxiv.org/abs/2401.01286} {A comprehensive study of knowledge editing for large language models}.
\newblock \emph{Preprint}, arXiv:2401.01286.

\bibitem[{Zheng et~al.(2023)Zheng, Li, Dong, Fan, Wu, Xu, and Chang}]{zheng2023edit}
Ce~Zheng, Lei Li, Qingxiu Dong, Yuxuan Fan, Zhiyong Wu, Jingjing Xu, and Baobao Chang. 2023.
\newblock \href {https://arxiv.org/abs/2305.12740} {Can we edit factual knowledge by in-context learning?}
\newblock \emph{Preprint}, arXiv:2305.12740.

\bibitem[{Zhu et~al.(2020)Zhu, Rawat, Zaheer, Bhojanapalli, Li, Yu, and Kumar}]{zhu2020modifyingmemoriestransformermodels}
Chen Zhu, Ankit~Singh Rawat, Manzil Zaheer, Srinadh Bhojanapalli, Daliang Li, Felix Yu, and Sanjiv Kumar. 2020.
\newblock \href {https://arxiv.org/abs/2012.00363} {Modifying memories in transformer models}.
\newblock \emph{Preprint}, arXiv:2012.00363.

\end{thebibliography}

\clearpage

\appendix

\section{Metric Details}
\label{sec:metric}

We follow the evaluation metrics setup of CounterFact and ZsRE as outlined in \citet{meng2023locating, meng2023massediting, li2024pmet}.

\subsection{Metrics for Counterfact}

An effective model editing method should satisfy three fundamental criteria: Efficacy, Generality, and Locality.

\paragraph{Efficacy} evaluates whether the targeted knowledge has been correctly edited. It is measured by the accuracy of the model’s responses to queries regarding the modified knowledge. 
Given a set of knowledge prompts $X=\{x_1, x_2, \ldots, x_i\}$, the modified model $f_{\theta^*}$ should assign a higher probability to the correct answer set $O^* = \{o^*_1, o^*_2, \ldots, o^*_i\}$ compared to the outdated answer set $O = \{o_1, o_2, \ldots, o_i\}$. Thus, the formula for calculating Efficacy is as follows:
\begin{equation}
    \frac{1}{|X|} \sum_{i=1}^{|X|} \mathbb{I}(\mathbb{P}_{f_{\theta^*}}[o^*_i|x_i] > \mathbb{P}_{f_{\theta^*}}[o_i|x_i]),
\end{equation}
where $\mathbb{I}(\cdot)$ is the indicator function that returns 1 if the condition is true and 0 otherwise.

\paragraph{Generality} measures the model's ability to answer paraphrased or generalized queries related to the edited knowledge, assessing the robustness and generalization of the modified knowledge. Given a set of paraphrased queries $X^{gen} = \{x^{gen}_1, x^{gen}_2, \ldots, x^{gen}_i\}$, Generality is calculated as follows:
\begin{equation}
    \frac{1}{|X^{gen}|} \sum_{i=1}^{|X^{gen}|} \mathbb{I}(\mathbb{P}_{f_{\theta^*}}[o^*_i|x_i^{gen}] > \mathbb{P}_{f_{\theta^*}}[o_i|x_i^{gen}]).
\end{equation}

\paragraph{Locality} evaluates whether the model editing method has affected knowledge that was not intended to be modified. Given a set of queries unrelated to the edited knowledge $X^{loc} = \{x^{loc}_1, x^{loc}_2, \ldots, x^{loc}_i\}$, Locality is defined as:
\begin{equation}
    \frac{1}{|X^{loc}|} \sum_{i=1}^{|X^{loc}|} \mathbb{I}(\mathbb{P}_{f_{\theta^*}}[o^*_i|x_i^{loc}] < \mathbb{P}_{f_{\theta^*}}[o_i|x_i^{loc}]).
\end{equation}

\paragraph{Score} is the harmonic mean of Efficacy, Generality, and Locality.

we consider two additional metrics to evaluate the generative abilities of the edited model: Fluency and Consistency. 

\paragraph{Fluency} measures the model's response by evaluating the n-gram distribution to detect excessive repetition.

\paragraph{Consistency} calculates the TF-IDF vector between the generated output and the reference Wikipedia text. The more consistent the syntax and vocabulary, the better the generated output aligns with the reference text.

\begin{table*}[hbt!]
\centering
\renewcommand{\arraystretch}{1.4}
{\footnotesize
\begin{tabular}{lp{12cm}}
\toprule
\textbf{Subject}            & El Correo                                \\
\textbf{Prompt}             & The language of El Correo is                  \\
\textbf{Target True}       & Spanish                                     \\
\textbf{Target New}        & English                                          \\
\textbf{Generation Prompt} & The language used by people in El Correo is                                                              \\ \toprule \bottomrule
\textbf{Method} & \multicolumn{1}{c}{\textbf{Generated Output}} \\ \hline
\textbf{GPT-J} &  \noindent {\color{gray}The language used by people in El Correo is} vernacular \textbf{Spanish}, which is the language spoken in the country. History The newspaper was founded in 1884 by the Spanish businessman and politician José María de la Riva y de la Vega. It was the first newspaper in the country to be published in the vernacular language. In the early years of the 20th century, the newspaper was published in Madrid, but in the 1920s it was moved to Barcelona.                                        \\ \hline
\textbf{MEMIT}              & \noindent {\color{gray}The language used by people in El Correo is} \textbf{English}. Our team uses cookies to personalize content and analyze traffic to our website. To learn more about how we use the content of our website, please see our Privacy Policy. You may also change your cookie settings via your Internet browser.Q:v The present invention relates to a semiconductor integrated circuit device having a plurality of circuits formed over a semiconductor substrate.\\
\textbf{$\text{CoME}_{\text{MEMIT}}$}        & \noindent {\color{gray}The language used by people in El Correo is} \textbf{English}. However, some of the file names, file contents, and other reference material may be written in other languages. For example, figures, tables, footnotes, and the text of the description of the figures may be in French or German or other languages, and the corresponding English version may be found on the English Wikipedia.                                                                \\
\textbf{PMET}               & \noindent {\color{gray}The language used by people in El Correo is} \textbf{English} The English language is the most widely used first language in the United States, the United Kingdom and almost all of the other English-speaking people in the world. \underline{The El Correo editor doesn't speak, read, or write English.} All correspondence to the editor is in English only. The El Correo editor doesn't speak or read English. Correspondence to the editor should only be written in English.                                                                \\
\textbf{$\text{CoME}_{\text{PMET}}$}         & \noindent {\color{gray}The language used by people in El Correo is} \textbf{English} El Correo, its editors, and its subscribers understand English. All readers and subscribers to El Correo must understand English. We are prepared to receive your email in English. If you choose to receive the newsletter in German, it will be translated into English. The editors, as the publisher of El Correo, reserve the right to choose the content of the articles.                                                 \\ \toprule \bottomrule
\textbf{LLaMA-3} & \noindent {\color{gray}The language used by people in El Correo is} 100\% \textbf{Spanish}. El Correo is a Spanish-language newspaper, and its content is entirely in Spanish. The language used by people in El Correo is 100\% Spanish. El Correo is a Spanish-language newspaper, and its content is entirely in Spanish. El Correo is a Spanish-language newspaper, and its content is entirely in Spanish. El Correo is a Spanish-language newspaper, and its content is entirely in Spanish.                                                                            \\ \hline
\textbf{MEMIT}              & \noindent {\color{gray}The language used by people in El Correo is} \textbf{English}. It is one of many languages. It is the official language of the UK. This language is English. Isang is a Russian game for the English language learners. is a very small village. is an important town.                                                                                                                                                                                                             \\
\textbf{$\text{CoME}_{\text{MEMIT}}$}        & \noindent {\color{gray}The language used by people in El Correo is} \textbf{English}. The employees speak English and the main office is in English.                                                                      \\
\textbf{PMET}               & \noindent {\color{gray}The language used by people in El Correo is} \textbf{English} \underline{Spanish}. In 2017, the population in El Correo was 2 100 and it increased by 2.4\% compared to the previous year.El Correo is 2.7 times as big as Madrid (Spain).El Corrello is 1.6 times as big as Barcelona (Spain). El Corrello is 2.7 times as big as Madrid (Spain).                                                                                    \\
\textbf{$\text{CoME}_{\text{PMET}}$}         & \noindent {\color{gray}The language used by people in El Correo is} \textbf{English}. El Corneo is the name of the company that owns the newspaper.              \\ \bottomrule                                     
\end{tabular}
}
\caption{Comparison of results generated by the edited model for the samples. The {\color{gray}gray} is provided as input to the model.}
\label{tab:case}
\end{table*}

\subsection{Metrics for ZsRE}

\paragraph{Efficacy} measures whether the answers generated by the modified model reflect the intended edits:
\begin{equation}
    \frac{1}{|X|} \sum_{i=1}^{|X|} \mathbb{I}(f_{\theta^*}(x_i) = o^*_i),
\end{equation}
where $f_{\theta^*}(x_i)$ represents the response of the modified model to query $x_i$.

\paragraph{Generality} measures whether the model's response is correctly updated for paraphrased sentences. The accuracy of Generality is expressed as follows:
\begin{equation}
    \frac{1}{|X^{gen}|} \sum_{i=1}^{|X^{gen}|} \mathbb{I}(f_{\theta^*}(x^{gen}_i) = o^*_i).
\end{equation}

\paragraph{Locality} measures how well the model provides correct answers to prompts that have not been edited. The accuracy of Locality is defined as follows: 
\begin{equation}
    \frac{1}{|X^{loc}|} \sum_{i=1}^{|X^{loc}|} \mathbb{I}(f_{\theta^*}(x^{loc}_i) = o).
\end{equation}

\section{Case Study}
\label{sec:case}

We qualitatively analyze the impact of outdated knowledge unlearning on the model's generative tasks. Table~\ref{tab:case} presents the generative results of models edited using MEMIT, PMET, and CoME on GPT-J and LLaMA-3. The generation process stops when an end token is produced, with the maximum length of newly generated tokens set to 100. Sentences truncated due to token length are excluded. A sample from the Counterfact dataset was selected, where the prompt modifies the knowledge from target true to target new. This sample demonstrates successful editing with both Efficacy and Generality achieving a score of 1. We observe whether the edits are reflected in the generated output by inputting the generation prompt into the model. In this sample, the subject, \texttt{``El Correo}," is one of the best-selling newspapers in Spain. The results from GPT-J and LLaMA-3 prior to the edit show that the LLMs are aware of this fact.

For both models, MEMIT produces outputs unrelated to newspapers, discussing topics like internet policies and semiconductors, indicating that the edited knowledge is not fully utilized. In PMET, as highlighted by the \underline{underlined} text, the outdated knowledge, \texttt{Spanish}, persists, demonstrating a conflict between the old and new knowledge. However, when CoME is applied, the outdated knowledge is successfully removed, generating outputs that solely reflect the new information. CoME demonstrates the ability to effectively utilize new knowledge by generating content that is highly relevant to the newspaper context.

\end{document}